\documentclass[sigconf]{acmart}

\usepackage{booktabs} 

\setcopyright{rightsretained}
\usepackage{graphicx}
\usepackage{amsmath}
\usepackage{amssymb}
\usepackage{enumitem}
\usepackage{multirow}
\usepackage{afterpage}
\usepackage{balance}

\copyrightyear{2018}
\acmYear{2018}
\setcopyright{acmlicensed}
\acmConference[MM '18]{2018 ACM Multimedia Conference}{October 22--26, 2018}{Seoul, Republic of Korea}
\acmBooktitle{2018 ACM Multimedia Conference (MM '18), October 22--26, 2018, Seoul, Republic of Korea}
\acmPrice{15.00}
\acmDOI{10.1145/3240508.3240587}
\acmISBN{978-1-4503-5665-7/18/10}


\begin{document}
\title[Generating Poetry from Images by Multi-Adversarial Training]{Beyond Narrative Description: Generating Poetry from Images by Multi-Adversarial Training}

\author{Bei Liu}
\authornote{This work was conducted when Bei Liu was a research intern at Microsoft Research.}
\affiliation{%
  \institution{Kyoto University}
}
\email{liubei@dl.kuis.kyoto-u.ac.jp}

\author{Jianlong Fu}
\authornote{Corresponding author}
\affiliation{%
  \institution{Microsoft Research Asia}
}
\email{jianf@microsoft.com}
\author{Makoto P. Kato}
\affiliation{%
  \institution{Kyoto University}
}
\email{mpkato@acm.org}
\author{Masatoshi Yoshikawa}
\affiliation{%
  \institution{Kyoto University}
}
\email{yoshikawa@i.kyoto-u.ac.jp}

\begin{abstract}
Automatic generation of natural language from images has attracted extensive attention. In this paper, we take one step further to investigate generation of poetic language (with multiple lines) to an image for automatic poetry creation. This task involves multiple challenges, including discovering poetic clues from the image (e.g., hope from green), and generating poems to satisfy both relevance to the image and poeticness in language level.
To solve the above challenges, we formulate the task of poem generation into two correlated sub-tasks by multi-adversarial training via policy gradient, through which the cross-modal relevance and poetic language style can be ensured. To extract poetic clues from images, we propose to learn a deep coupled visual-poetic embedding, in which the poetic representation from objects, sentiments \footnote{We consider both adjectives and verbs that can express emotions and feelings as sentiment words in this research.} and scenes in an image can be jointly learned. Two discriminative networks are further introduced to guide the poem generation, including a multi-modal discriminator and a poem-style discriminator. To facilitate the research, we have released two poem datasets by human annotators with two distinct properties: 1) the first human annotated image-to-poem pair dataset (with $8,292$ pairs in total), and 2) to-date the largest public English poem corpus dataset (with $92,265$ different poems in total). Extensive experiments are conducted with 8K images, among which 1.5K image are randomly picked for evaluation. Both objective and subjective evaluations show the superior performances against the state-of-the-art methods for poem generation from images. Turing test carried out with over $500$ human subjects, among which 30 evaluators are poetry experts, demonstrates the effectiveness of our approach.
\end{abstract}


\maketitle

\section{Introduction} \label{introduction}

Researches that involve both vision and languages have attracted great attentions recently as we can witness from the bursting works on image descriptions like image caption and paragraph \cite{chen2017show, fang2015captions, krause2016hierarchical, vinyals2015show}. Image descriptions aim to generate sentence(s) to describe facts from images in human-level languages. In this paper, we take one step further to tackle a more cognitive task: generation of poetic language to an image for the purpose of poetry creation, which has attracted tremendous interest in both research and industry fields.

\begin{figure}[t]
\begin{center}
\includegraphics[width=0.9\linewidth]{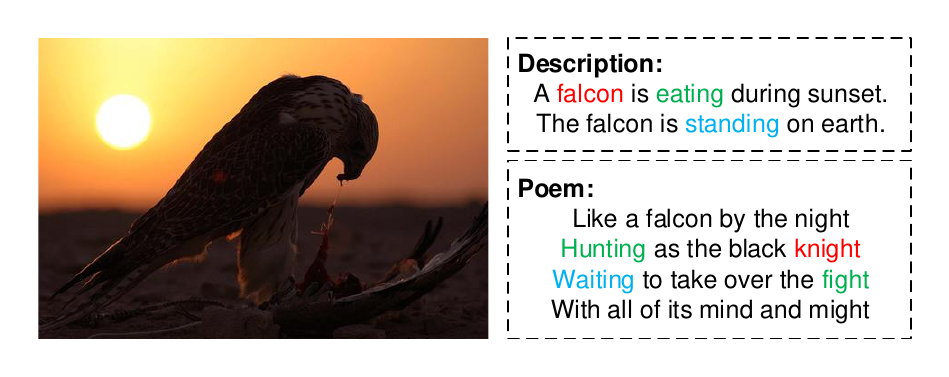}
\end{center}
\vspace{-4mm}
\caption{\small{Example of human written description and poem of the same image. We can see a significant difference from words of the same color in these two forms. Instead of describing facts in the image, poem tends to capture deeper meaning and poetic symbols from objects, scenes and sentiments from the image (such as \textbf{knight} from \textit{falcon}, \textbf{hunting} and \textbf{fight} from \textit{eating}, and \textbf{waiting} from \textit{standing}). }}
\label{fig:example}
\vspace{-6mm}
\end{figure}

In natural language processing field, poem generation related problems have been studied. In \cite{hopkins2017automatically, yan2013poet}, the authors mainly focused on the quality of style and rhythm. In \cite{ghazvininejad2016generating, yan2013poet, zhang2014chinese}, these works have taken one more step to generate poems from topics. Image inspired Chinese quatrain generation is proposed in \cite{xu2018images}. In the industrial field, Facebook has proposed to generate English rhythmic poetry with neural networks \cite{hopkins2017automatically}, and Microsoft has developed a system called XiaoIce, in which poem generation is one of the most important features. Nevertheless, generating poems from images in an end-to-end fashion remains a new topic with grand challenges.

Compared with image captioning and paragraphing that focus on generating descriptive sentences about an image, generation of poetic language is a more challenging problem. There is a larger gap between visual representations and poetic symbols that can be inspired from images and facilitate better generation of poems. For example, ``man'' detected in image captioning can further indicate ``hope'' with ``bright sunshine'' and ``opening arm'', or ``loneliness'' with ``empty chairs'' and ``dark'' background in poem creation. Fig.~(\ref{fig:example}) shows a concrete example of the differences between descriptions and poems for the same image.

In particular, to generate a poem from an image, we are facing with the following three challenges.
First of all, it is a cross-modality problem compared with poem generation from topics. An intuitive way for poem generation from images is to first extract keywords or captions from images and then consider them as seeds for poem generation as what poem generation from topics do. However, keywords or captions will miss a lot of information in images, not to mention the poetic clues that are important for poem generation \cite{ghazvininejad2016generating,zhang2014chinese}.
Secondly, compared with image captioning and image paragraphing, poem generation from images is a more subjective task, which means an image can be relevant to several poems from various aspects while image captioning/paragraphing is more about describing facts in the images and results in similar sentences. Thirdly, the form and style of poem sentences is different from that of narrative sentences. In this research, we mainly focus on \textit{free verse} which is an open form of poetry. Although we do not require meter, rhyme or other traditional poetic techniques, it remains some sense of poetic structures and poetic style language in poems. We define this quality of poem as \textit{poeticness} in this research. For example, length of poems are usually not very long, specific words are preferred in poems compared with image descriptions, and sentences in one poem should be consistent to one topic.

To address the above challenges, we collect two poem datasets by human annotators, and propose poetry creation by integrating retrieval and generation techniques in one system. Specifically, to better learn poetic clues from images for poem generation, we first learn a deep coupled visual-poetic embedding model with CNN features of images, and skip-thought vector features \cite{kiros2015skip} of poems from a multi-modal poem dataset (namely ``MultiM-Poem'') that consists of thousands of image-poem pairs. This embedding model is then used to retrieve relevant and diverse poems from a larger uni-modal poem corpus (namely ``UniM-Poem'') for images. Images with these retrieved poems and MultiM-Poem together construct an enlarged image-poem pair dataset (namely ``MultiM-Poem (Ex)''). We further propose to leverage the state-of-art sequential learning techniques for training an end-to-end image to poem model on the MultiM-Poem (Ex) dataset. Such a framework ensures substantial poetic clues, that are significant for poem generation, could be discovered and modeled from those extended pairs.

To avoid exposure bias problems caused by long length of long sequence (all poem lines together) and the problem that there is no specific loss available to score a generated poem,
we propose to use a recurrent neural network (RNN) for poem generation with multi-adversarial training and further optimize it by policy gradient. Two discriminative networks are used to provide rewards in terms of the generated poem's relevance to the given image and poeticness of the generated poem.
We conduct experiments on MultiM-Poem, UniM-Poem and MultiM-Poem (Ex) to generate poems to images. The generated poems are evaluated in both objective and subjective ways. We define automatic evaluation metrics concerning relevance, novelty and translative consistence and conducted user studies about relevance, coherence and imaginativeness of generated poems to compare our model with baseline methods.
The contributions in this research are concluded as follows:
\begin{itemize} [nosep]
\item We propose to generate poems (English free verse) from images in an end-to-end fashion. To the best of our knowledge, this is the first attempt to study the image-inspired English poem generation problem in a holistic framework, which enables a machine to approach human capability in cognition tasks.
\item We incorporate a deep coupled visual-poetic embedding model and a RNN-based generator for joint learning, in which two discriminators provide rewards for measuring cross-modality relevance and poeticness by multi-adversarial training.
\item We collect the first paired dataset of image and poem annotated by human annotators, and the largest public poem corpus dataset. Extensive experiments demonstrate the effectiveness of our approach compared with several baselines by using both objective and subjective evaluation metrics, including a Turing test from more than $500$ human subjects. To better promote the research in poetry generation from images, we have released these datasets on Github\footnote{\url{https://github.com/bei21/img2poem}}.
\end{itemize}

\begin{figure*}[th]
\vspace{-9mm}
\begin{center}
\includegraphics[width=0.9\linewidth]{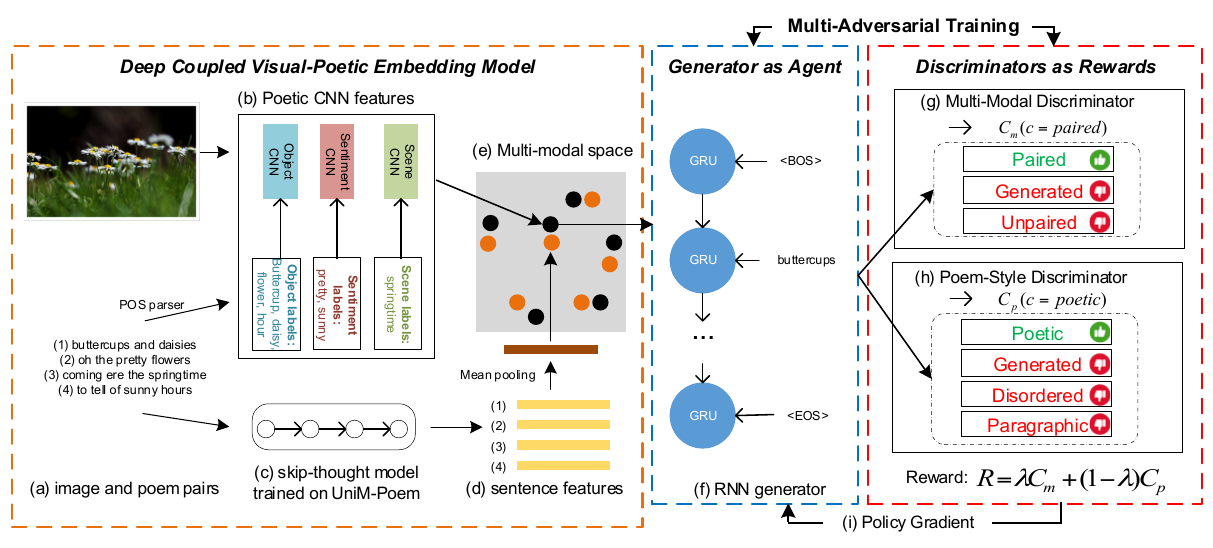}
\end{center}
\vspace{-6mm}
\caption{\small{The framework of poetry generation with multi-adversarial training. A deep coupled visual-poetic model (e) is trained by human annotated image-poem pairs (a). The image features (b) are poetic multi-CNN features obtained by fine-tuning CNNs with the extracted poetic symbols (e.g., objects, scenes and sentiments) by a POS parser \cite{toutanova2003feature} from poems. The sentence features (d) of poems are extracted from a skip-thought model (c) trained on the largest public poem corpus (UniM-Poem). A RNN-based sentence generator (f) is trained as agent and two discriminators considering multi-modal (g) and poem-style (h) critics of a generated poem to a given image provide rewards to policy gradient (i). POS parser extracts Part-Of-Speech words from poems. }}
\label{fig:overall}
\vspace{-4mm}
\end{figure*}

\section{Related Work}

\subsection{Poetry Generation}

Traditional approaches for poetry generation include template and grammar-based method \cite{manurung1999chart, oliveira2009automatic, oliveira2012poetryme}, generative summarization under constrained optimization \cite{yan2013poet} and statistical machine translation model \cite{he2012generating, jiang2008generating}. By applying deep learning approaches recent years, researches about poetry generation has entered a new stage. Recurrent neural network is widely used to generate poems that can even confuse readers from telling them from poems written by human poets \cite{ghazvininejad2016generating, ghazvininejad2017hafez, hopkins2017automatically, yi2017generating, zhang2014chinese}.
 Previous works of poem generation mainly focus on style and rhythmic qualities of poems \cite{hopkins2017automatically, yan2013poet}, while recent studies introduce topic as a condition for poem generation \cite{ghazvininejad2016generating, ghazvininejad2017hafez, yan2013poet, zhang2014chinese}.
 For a poem, topic is still a rather abstract concept without specific scenarios.
 Inspired by the fact that many poems were created in a conditioned scenario, we take one step further to tackle the problem of generating poems inspired by a visual scenario. Compared with previous researches, our work is facing with more challenges, especially in terms of multi-modal problems.

\subsection{Image Description}

Image captioning is first regarded as a retrieval problem which aims to search captions from dataset for a given image \cite{farhadi2010every, karpathy2014deep} and hence cannot provide accurate and proper descriptions for all images. To overcome this problem, methods like template filling \cite{kulkarni2011baby} and paradigm for integrating convolutional neural network (CNN) and recurrent neural network (RNN) \cite{chen2015mind, vinyals2015show, you2016image, yao2017boosting} are proposed to generate readable human-level sentences. Recently, generative adversarial network (GAN) is applied to generate captions based on different problem settings \cite{chen2017show, yu2017seqgan}. Similarly to image captioning, image paragraphing is going the similar way. Recent researches about image paragraphing mainly focus on region detection and hierarchical structure for generated sentences \cite{krause2016hierarchical, liu2017let, park2015expressing}.
However, as we have addressed, image captioning and paragraphing aim to generate descriptive sentences to tell the facts in images, while poem generation is tackling an advanced form of linguistic form which requires poeticness and language style constrains.

\section{Approach}

In this research, we aim to generate poems from images so that the generated poems are relevant to input images and satisfy poeticness.
For this purpose, we cast our problem in a multi-adversarial procedure \cite{goodfellow2014generative} and further optimize it with a policy gradient \cite{williams1992simple, zaremba2015reinforcement}. A CNN-RNN generative model acts as an \textit{agent}. The parameters of this agent define a \textit{policy} whose execution will decide which word to be picked as an \textit{action}. When the agent has picked all words in a poem, it observes a \textit{reward}. We define two discriminative networks to serve as rewards concerning whether the generated poem is a paired one with the input image and whether the generated poem is poetic. The goal of our poem generation model is to generate a sequence of words as a poem for an image to maximize the expected end reward. This policy-gradient method has shown significant effectiveness to many tasks without non-differentiable metrics \cite{chen2017show, rennie2017self, yu2017seqgan}.

As shown in Fig.~(\ref{fig:overall}), the framework consists of several parts: (1) a deep coupled visual-poetic embedding model to learn poetic representations from images, and (2) a multi-adversarial training procedure optimized by policy gradient. A RNN based generator serves as agent, and two discriminative networks provide rewards to the policy gradient.

\subsection{Deep Coupled Visual-Poetic Embedding} \label{section:embedding}

The goal of visual-poetic embedding model \cite{frome2013devise, kiros2014unifying} is to learn an embedding space where points of different modality, e.g. images and sentences, can be projected to. In a similar way to image captioning problem, we assume that a pair of image and poem shares similar poetic semantics which makes the embedding space learnable. By embedding both images and poems to the same feature space, we can directly compute the relevance between a poem and an image by poetic vector representations of them. Moreover, the embedding feature can be further utilized to initialize a optimized representation of poetic clues for poem generation.

The structure of our deep coupled visual-poetic embedding model is shown in left part of Fig.~(\ref{fig:overall}). For the input of images, we leverage three deep convolutional neural networks (CNNs) concerning three aspects that indicate important poetic clues from images inspired from fine-grained problems \cite{LookCloser}, namely object ($\mathbf{v}_1$), scene ($\mathbf{v}_2$) and sentiment ($\mathbf{v}_3$), after conducting a prior user study about important factors for poem creation from images. We observed that concepts in poems are often imaginative and poetic while concepts in the classification datasets we use to train our CNN models are concrete and common. To narrow the semantic gap between the visual representation of images and the textual representation of poems, we propose to fine-tune these three networks with MultiM-Poem dataset. Specifically, frequent used keywords about object, sentiment and scenes in the poems are picked as label vocabulary, and then we build three multi-label datasets based on MultiM-Poem dataset for object, sentiment and scenes detection respectively. Once the multi-label datasets are built, we fine-tune the pre-trained CNN models on the three datasets independently, which is optimized by sigmoid cross entropy loss as shown in Eq.~\eqref{eqn:loss}. After that, we adopt the $D$-dimension deep features for each aspect from the penultimate fully-connected layer of the CNN models, and get a concatenated $N$-dimension ($N=D\times3$) feature vector $\mathbf{v}\in \mathbb{R}^{N}$ as input of visual-poetic embedding for each image:
\vspace{-2mm}
\begin{eqnarray}
\label{eqn:loss}
loss = \frac{-1}{N}\sum_{n=1}^{N}(t_nlogp_n + (1-t_n)log(1-p_n)),\\
\vspace{-2mm}
\begin{split}
\mathbf{v}_1=f_{\mathrm{Object}}(I),&&
\mathbf{v}_2=f_{\mathrm{Scene}}(I),\\
\mathbf{v}_3=f_{\mathrm{Sentiment}}(I),&&
\mathbf{v}=(\mathbf{v}_1, \mathbf{v}_2, \mathbf{v}_3).
\end{split}
\vspace{-5mm}
\end{eqnarray}
The output of visual-poetic embedding vector $\mathbf{x}$ is a $K$-dimension vector representing the image embedding with linear mapping from image features:
\vspace{-1mm}
\begin{equation}
\mathbf{x}=\mathbf{W}_{v}\cdot \mathbf{v} + \mathbf{b}_{v}\in \mathbb{R}^{K},
\end{equation}
where $\mathbf{W}_{v} \in \mathbb{R}^{K\times N}$ is the image embedding matrix and $\mathbf{b}_{v}\in \mathbb{R}^{K}$ is the image bias vector.
Meanwhile, representation feature vector of a poem is computed by skip-thought vectors\cite{kiros2015skip}, which is a popular unsupervised method to learn sentence embedding. We train skip-thought model on unpaired UniM-Poem dataset and use it to provide a better sentence representation for poem sentences.
Mean value of all sentences' combined skip-thought features (unidirectional and bidirectional) is denoted by $\mathbf{t}\in\mathbb{R}^{M}$ where $M$ is the combined dimension. Similar to image embedding, the poem embedding is denoted as:
\vspace{-1mm}
\begin{equation}
\mathbf{m}=\mathbf{W}_{t}\cdot \mathbf{t} + \mathbf{b}_{t}\in \mathbb{R}^{K},
\vspace{-1mm}
\end{equation}
where $\mathbf{W}_{t} \in \mathbb{R}^{K\times M}$ for the poem embedding matrix and $\mathbf{b}_{t}\in \mathbb{R}^{K}$ for the poem bias vector.
Finally, the image and poem are embedded together by minimizing a pairwise ranking loss with dot-product similarity:
\vspace{-2mm}
\begin{equation}
\begin{split}
L=\sum_{\mathbf{x}}\sum_{k}\mathrm{max}(0,\alpha -\mathbf{x}\cdot\mathbf{m}+\mathbf{x}\cdot\mathbf{m}_{k})  \\
+\sum_{\mathbf{m}}\sum_{k}\mathrm{max}(0,\alpha -\mathbf{m}\cdot\mathbf{x}+\mathbf{m}\cdot\mathbf{x}_{k}),
\end{split}
\label{eq:embedding}
\vspace{-8mm}
\end{equation}
where $\mathbf{m}_{k}$ is a contrastive (irrelevant unpaired) poem for image embedding $\mathbf{x}$, and vice-versa with $\mathbf{x}_{k}$. $\alpha$ denotes the contrastive margin.
As a result, the model we trained will produce higher dot-product similarity between embedding features of image-poem pairs than similarity between randomly generated pairs.

\subsection{Poem Generator as an Agent}

A conventional CNN-RNN model for image captioning is used to serve as an agent.
Instead of using hierarchical methods that are used recently in generating multiple sentences \cite{krause2016hierarchical}, we use a non-hierarchical recurrent model by treating the end-of-sentence token as a word in the vocabulary. The reason is that 1) poems often consist of fewer words compared with paragraphs; 2) there is lower consistent hierarchy between sentences of poems, which makes the hierarchy much more difficult to learn. We also conduct experiment with hierarchical recurrent language model as a baseline and we will show the result in the experiment part.

The generative model includes CNNs for image encoder and a RNN for poem decoder.  The reason of using RNN instead of CNN  for languages is that it can better encode the structure-dependent semantics of the long sentences which are widely observed in poems. In this research, we apply Gated Recurrent Units (GRUs) \cite{chung2014empirical} for poem decoder for its simple structure and robustness to overfitting problem on less training data. We use image-embedding features learned by the deep coupled visual-poetic embedding model explained in Section \ref{section:embedding} as input of image encoder. Suppose $\theta$ is the parameters of the model. Traditionally, our target is to learn $\theta$ by maximizing the likelihood of the observed sentence $\mathbf{y}=y_{1:T}\in\mathbb{Y^*}$ where $T$ is the maximum length of generated sentence (including $<\textrm{BOS}>$ for start of sentence, $<\textrm{EOS}>$ for end of sentence and line breaks) and $\mathbb{Y^*}$ denotes a space of all sequences of selected words.

Let $r(y_{1:t})$ denote the reward achieved at time $t$ and $R(y_{1:\tiny{T}})$ is the cumulative reward, namely $R(y_{k:T})=\sum_{t=k}^{T}r(y_{1:t})$. Let $p_{\theta}(y_{t}|y_{1:(t-1)})$ be a parametric conditional probability of selecting $y_{t}$ at time step $t$ given all the previous words $y_{1:(t-1)}$. $p_\theta$ is defined as a parametric function of policy $\theta$. The reward of policy gradient in each batch can be computed as the sum over all sequences of valid actions as the expected future reward. To iterate over sequences of all possible actions is exponential, but we can further write it in expectation so that it can be approximated with an unbiased estimator:
\vspace{-2mm}
\begin{equation}
\begin{split}
J(\theta)=\sum_{y_{1:T}\in \mathbb{Y^*}}p_\theta(y_{1:T})R(y_{1:T})
=\mathbb{E}_{y_{1:T}\sim p_\theta}\sum_{t=1}^T r(y_{1:t}).
\end{split}
\vspace{-4mm}
\end{equation}
We aim to maximize $J(\theta)$ by following its gradient:
\vspace{-2mm}
\begin{equation}
\nabla_\theta J(\theta)=\mathbb{E}_{y_{1:T}\sim p_\theta}\left [\sum_{t=1}^T \nabla_\theta \mathrm{log}p_\theta(y_{1:t-1})\right ]\sum_{t=1}^T r(y_{1:t}).
\vspace{-2mm}
\end{equation}
In practice the expected gradient can be approximated using a Monte-Cartlo sample by sequentially sample each $y_t$ from the model distribution $p_\theta(y_t|y_{1:(t-1)})$ for $t$ from $1$ to $T$.
As discussed in \cite{rennie2017self}, a baseline $b$ can be introduce to reduce the variance of the gradient estimate without changing the expected gradient. Thus, the expected gradient with a single sample is approximated as follow:
\vspace{-2mm}
\begin{equation}
\nabla_\theta J(\theta) \approx \sum_{t=1}^T \nabla_\theta \mathrm{log} p_\theta(y_{1:t-1}) \sum_{t=1}^T (r(y_{1:t})-b_t).
\vspace{-1mm}
\end{equation}

\subsection{Discriminators as Rewards}

A good poem for an image has to satisfy at least two criteria: the poem (1) is relevant to the image, and (2) has some sense of poectiness concerning proper length, poem's language style and consistence between sentences. Based on these two requirements, we propose two discriminative networks to guide the generated poem: multi-modal discriminator and poem-style discriminator.

\textbf{Multi-Modal Discriminator.}
Multi-modal discriminator ($D_{m}$) is used to guide the generated poem $\mathbf{y}$ related to corresponding image $\mathbf{x}$. It is trained to classify a poem into three classes: \textit{paired} as positive examples, \textit{unpaired} and \textit{generated} as negative examples. \textit{Paired} includes ground-truth paired poems for the input images. \textit{Unpaired} poems are randomly sampled from unpaired poems of the input images in training data.
$D_{m}$ includes a multi-modal encoder, modality fusion layer and a classifier with softmax function:
\vspace{-2mm}
\begin{eqnarray}
\mathbf{c}=\mathrm{GRU}_\rho(\mathbf{y}), \\
f=\mathrm{tanh}(W_x \cdot \mathbf{x} + b_x)\odot  \mathrm{tanh}(W_c \cdot \mathbf{c} + b_c), \\
C_m=\mathrm{softmax}(W_m \cdot f + b_m), \label{mm_softmax}
\vspace{-2mm}
\end{eqnarray}
where $\rho$, $W_x$, $b_x$, $W_c$, $b_c$, $W_m$, $b_m$ are parameters to be learned, $\odot$ is element-wise multiplication and $C_m$ denotes the probabilities over three classes of the multi-modal discriminator. We utilize GRU-based sentence encoder for discriminator training.  Eq.~(\ref{mm_softmax}) provides way to generate the probability of $(\mathbf{x},\mathbf{y}$ classified into each class as denoted by $C_m(c|\mathbf{x},\mathbf{y})$ where $c\in\{\textit{paired}, \textit{unpaired}, \textit{generated}\}$.

\textbf{Poem-Style Discriminator.}
In contrast with most poem generation researches that emphasize on meter, rhyme or other traditional poetic techniques, we focus on \textit{free verse} which is an open form of poetry. Even though, we require our generated poems have the quality of \textit{poeticness} as we define in Section \ref{introduction}. Without making specific templates or rules for poems, we propose a poem-style discriminator ($D_p$) to guide generated poems towards human written poems.
In $D_p$, generated poems will be classified into four classes: \textit{poetic}, \textit{disordered}, \textit{paragraphic} and \textit{generated}.

Class \textit{poetic} is addressed as positive example of poems that satisfy poeticness. The other three classes are all regarded as negative examples. Class \textit{disordered} concerns about the inner structure and coherence between sentences of poems and \textit{paragraphic} class uses paragraph sentences as negative examples.
In $D_p$, we use UniM-Poem as positive \textit{poetic} samples. To construct disordered poems, we first construct a poem sentence pool by splitting all poems in UniM-Poem. Examples of class \textit{disordered} are poems that we reconstruct by sentences randomly picked up with a reasonable line numbers from poem sentence pool. Paragraph dataset provided by \cite{krause2016hierarchical} is used as \textit{paragraph} examples.

A completed generated poem $\mathbf{y}$ is encoded by GRU and parsed to a fully connected layer, and the probability of falling into four classes is computed by a softmax function. Formula of this procedure is as follow:
\vspace{-1mm}
\begin{equation}
C_p=\mathrm{softmax}(W_p \cdot \mathrm{GRU}_\eta (\mathbf{y}) + b_p), \label{p_softmax}
\vspace{-1mm}
\end{equation}
where $\eta$, $W_p$, $b_p$ are parameters to be learned. The probability of classifying generated poem $\mathbf{y}$ to a class $c$ is formulated as $C_p(c|\mathbf{y})$ where $c\in\{\textit{poetic}, \textit{disordered}, \textit{paragraphic}, \textit{generated}\}$.

\textbf{Reward Function.} We define the reward function for policy gradient as a linear combination of probability of classifying generated poem $\mathbf{y}$ for an input image $\mathbf{x}$ to the positive class (\textit{paired} for multi-modal discriminator $D_m$ and \textit{poetic} for poem-style discriminator $D_p$) weighted by tradeoff parameter $\lambda$:
\vspace{-1mm}
\begin{equation}
R(\mathbf{y}|\cdot )=\lambda  C_m(c=\textit{paired}|\mathbf{x},\mathbf{y}) + (1-\lambda)C_p(c=\textit{poetic}|\mathbf{y}).
\label{equ:objective}
\vspace{-3mm}
\end{equation}

\subsection{Multi-Adversarial Training}

Adversarial training is a minimax game between a generator $G$ and a discriminator $D$ with value function $V(G,D)$:
\vspace{-1mm}
\begin{equation}
\scriptstyle{\underset{G}{\textrm{min}}\underset{D}{\textrm{max}}V(D,G)=\mathbb{E}_{x\sim p_\textrm{data}(x)}[\textrm{log}D(x)]+\mathbb{E}_{z\sim p_z(z)}[\textrm{log}(1-D(G(z)))].}
\vspace{-1mm}
\end{equation}
We propose to use multiple discriminators by reformulating $G$'s objective as:
\vspace{-1mm}
\begin{equation}
\underset{G}{\textrm{min}}\textrm{max}F(V(D_1,G),...,V(D_n,G)),
\vspace{-1mm}
\end{equation}
where we have $n=2$, and $F$ indicates linear combination of discriminators as shown in Eq.~\eqref{equ:objective}.

The generator aims to generate poems that have higher rewards for both discriminators so that they can fool the discriminators while the discriminators are trained to distinguish the generated poems from \textit{paired} and \textit{poetic} poems. The probabilities of classifying generated poem into positive classes in both discriminators are used as rewards to policy gradient as explained above.

Multiple discriminators (two in this work) are trained by providing positive examples from the real data (paired poems in $D_m$ and poem corpus in $D_p$) and negative examples from poems generated from the generator as well as other negative forms of real data (unpaired poems in $D_m$, paragraphs and disordered poems in $D_p$.
Meanwhile, by employing a policy gradient and Monte Carlo sampling,  the generator is updated based on the expected rewards from multiple discriminators.
Since we have two discriminators, we apply a multi-adversarial training method that will train two discriminators in a parallel way.

\section{Experiments}

\subsection{Datasets}

\begin{figure}[t]
\begin{center}
\includegraphics[width=1\linewidth]{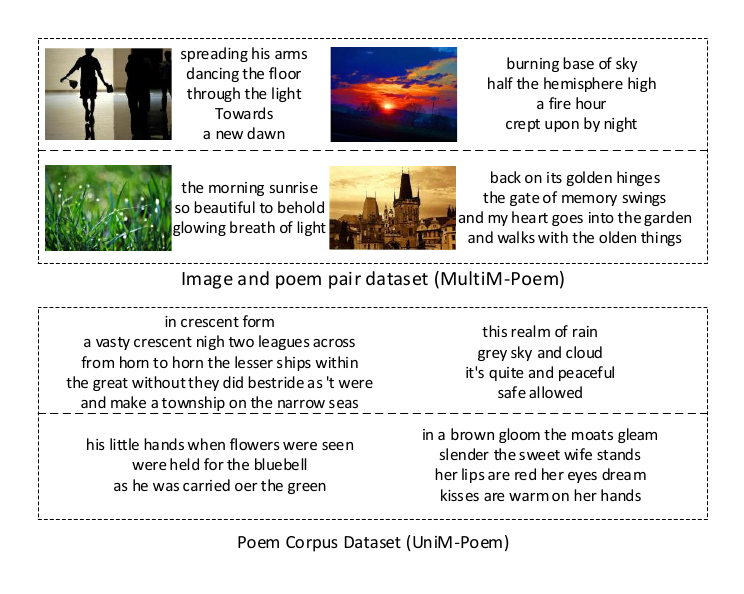}
\end{center}
\vspace{-6mm}
\caption{Examples in two datasets: UniM-Poem and MultiM-Poem. }
\label{fig:dataset}
\vspace{-4mm}
\end{figure}

\begin{table}[h]\small
\begin{center}
\begin{tabular}{cccccc}
\hline
 Name & \#Poem & \#Line/poem &\#Word/line\\\hline\hline
MultiM-Poem & 8,292  & 7.2 & 5.7\\ \hline
UniM-Poem & 93,265  & 5.7 & 6.2\\ \hline\hline
 MultiM-Poem (Ex) & 26,161 & 5.4 & 5.9 \\
\hline
\end{tabular}
\end{center}
\caption{\small{Detailed information about the three datasets. The first two datasets are collected by ourselves and the third one is extended by our embedding model.}}
\label{table:dataset}
\vspace{-8mm}
\end{table}

To facilitate the research of poetry generation from images, we collected two poem datasets, in which one consists of image and poem pairs, namely Multi-Modal Poem dataset (MultiM-Poem), and the other is a large poem corpus, namely Uni-Modal Poem dataset (UniM-Poem). By using the embedding model we have trained, the image and poem pairs are extended by adding the nearest three neighbor poems from the poem corpus without redundancy, and an extended image and poem pair dataset is constructed and denoted as MultiM-Poem (Ex). The detailed information about these datasets is listed in Table \ref{table:dataset}.
Examples of the two collected datasets can be seen in Fig. \ref{fig:dataset}.

For MultiM-Poem dataset, we first crawled 34,847 image-poem pairs in Flickr from groups that aim to use images illustrating poems written by human. Five human assessors majoring in English literature were further asked to evaluate these poems as relevant or irrelevant by judging whether the image can exactly inspire the poem in a pair by considering the associations of objects, sentiments and scenes. We filtered out pairs labeled as irrelevant and kept the remaining 8,292 pairs to construct the MultiM-Poem dataset.

UniM-Poem is crawled from several public online poetry websites, such as Poetry Foundation\footnote{\url{https://www.poetryfoundation.org/}}, PoetrySoup\footnote{\url{https://www.poetrysoup.com/}}, best-poem.net and poets.org. To achieve robust model training, a poem pre-processing procedure is conducted to filter out those poems with too many lines ($>10$) or too fewer lines ($<3$). We also remove poems with strange characters, poems in languages other than English and duplicate poems.

\begin{figure*}
\vspace{-9mm}
\centering
\includegraphics[width=0.99\linewidth]{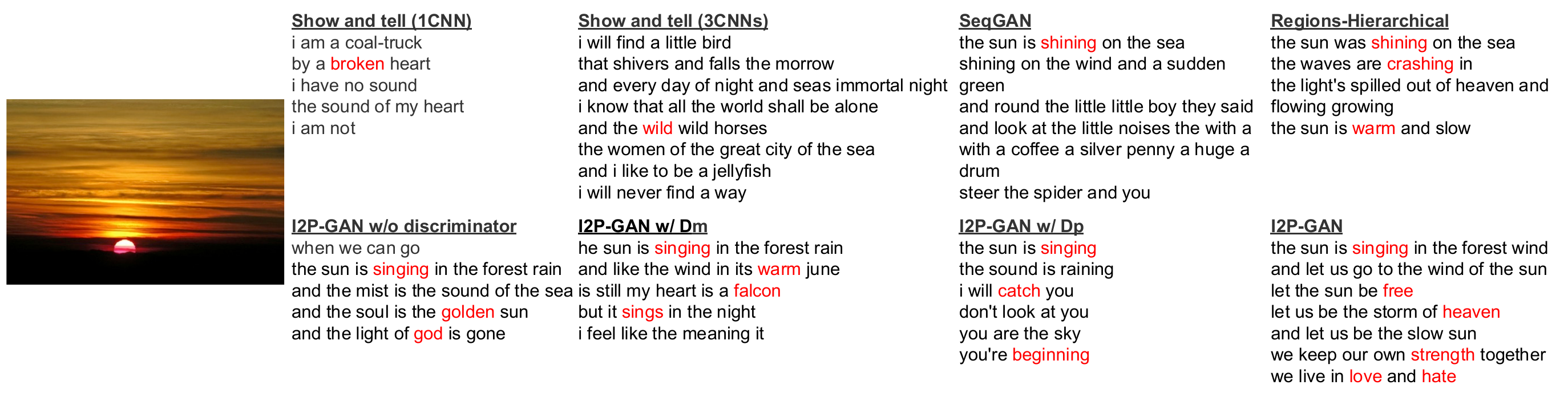}
\vspace{-5mm}
\caption{\small{Example of poems generated by eight methods for an image. Words in read indicate poeticness.}}
\label{fig:comparedexample}
\vspace{-4mm}
\end{figure*}

\begin{figure*}
\centering
\includegraphics[width=0.99\linewidth]{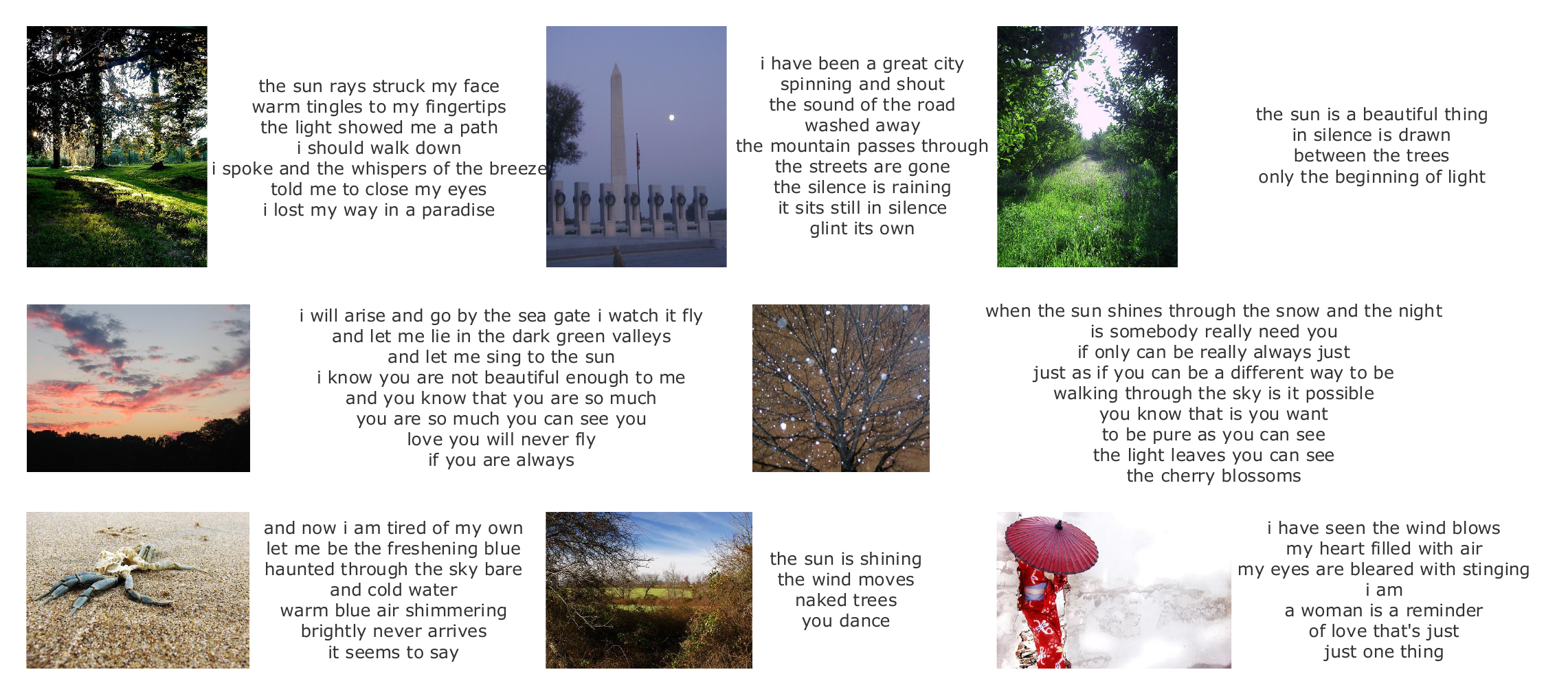}
\vspace{-3mm}
\caption{Example of poems generated by our approach I2P-GAN.}
\label{fig:generatedexample}
\vspace{-4mm}
\end{figure*}

\subsection{Compared Methods}

To investigate the effectiveness of the proposed methods, we compare with four baseline models with different settings. The models of show-and-tell \cite{vinyals2015show} and SeqGan \cite{yu2017seqgan} are selected due to their state-of-art results in image captioning. A competitive image paragraphing model is selected, as its strong capability for modeling diverse image content. Note that all the methods use MultiM-Poem (Ex) as the training dataset, and can generate multiple lines as poems. The detailed experiment settings are shown as follows:

\textbf{Show and tell (1CNN):} CNN-RNN model trained with only object CNN by VGG-16 .

\textbf{Show and tell (3CNNs):} CNN-RNN model trained with three CNN features by VGG-16.

\textbf{SeqGAN:} CNN-RNN model optimized with a discriminator to tell from generated poems and ground-truth poems. We use RNN for discriminator for fair comparison.

\textbf{Regions-Hierarchical:} Hierarchical paragraph generation model based on \cite{krause2016hierarchical}. To better align with poem distribution, we restrict the maximum lines to be 10 and each line has up to 10 words in the experiment.

\textbf{Our Model:} To demonstrate the effectiveness of the two discriminators, we train our model (Image to Poem with GAN, I2P-GAN) in four settings: pretrained model without discriminators (\textbf{I2P-GAN w/o discriminator}), with multi-modal discriminator only (\textbf{I2P-GAN w/ $\mathbf{D_m}$}), with poem-style discriminator only (\textbf{I2P-GAN w/ $\mathbf{D_p}$}) and with both discriminators (\textbf{I2P-GAN}).

\subsection{Automatic Evaluation Metrics}

Evaluation of poems is generally a difficult task and there are no established metrics in existing works, not to mention the new task of generating poems from images. To better address the performance of the generated poems, we propose to evaluate them in both automatic and manual way.

We propose to employ three metrics for automatic evaluation, e.g., BLEU, novelty and relevance. An overall score is computed by the three metrics after normalization.

\textbf{BLEU.}
We use Bilingual Evaluation Understudy (BLEU) \cite{papineni2002bleu} score-based evaluation to examine how likely the generated poems can approximate towards the ground-truth ones following image captioning and paragraphing. It is also used in some poem generation works \cite{yan2013poet}. For each image, we only use the human written poems as ground-truth poems.

\textbf{Novelty.}
By introducing discriminator $D_p$, the generator is supposed to introduce words or phrases from UniM-Poem dataset and results in words or phrases that are not very frequent in MultiM-Poem (Ex) dataset. We use \textit{novelty} as proposed by \cite{xu2017neural} to measure the number of infrequent words or phrases observed in the generated poems. Two scales of N-gram are explored, e.g. bigram and trigram, as \textbf{Novelty-2} and \textbf{Novelty-3}. We first rank the n-grams that occur in the training dataset of MultiM-Poem (Ex) and take the top 2,000 as frequent ones.
Novelty is computed as the proportion of n-grams that occur in training dataset except the frequent ones in the generated poem.

\textbf{Relevance.}
Different from poem generation researches that have no or weak constrains to poem contents, we consider relevance of the generated poem to the given image as an important measurement in this research. However, unlike captions that concern more about facts about images, different poems can be relevant to the same image from various aspects. Thus, instead of computing relevance between generated poem and ground-truth poems, we define relevance between a poem and an image using our learned deep coupled visual-poetic embedding (VPE) model. After mapping the image and the poem to the same space through VPE, linearly scaled cosine similarity (0-1) is used to measure their relevance.

\textbf{Overall.}
We compute an overall score based on the above three metrics. For each value $a_i$ in all values of one metric $\mathbf{a}$, we first linearly normalize it with following method:
\vspace{-2mm}
\begin{equation}
{a_i}'=\frac{a_i-\mathrm{min}(a)}{\mathrm{max}(a)-\mathrm{min}(a)}.
\label{metric:normal}
\vspace{-2mm}
\end{equation}
After that, we get average values for BLEU (e.g. BLEU-1, BLEU-2 and BLEU-3) and novelty (e.g. Novelty-2 and Novelty-3). A final score is computed by averaging the normalized values, to ensure equal contribution of different metrics.

However, in such an open-ended task, there are no particularly suitable metrics that can perfectly evaluate the performance of generated poems. The automatic metrics we use can be used as a guidance to some extent. To better illustrate the performance of poems from human perception, we further conduct extensive user studies in the follows.

\subsection{Human Evaluation}
We conducted human evaluation in Amazon Mechanical Turk. In particular, three types of tasks are assigned:

\textbf{Task1}: to explore the effectiveness of our deep coupled visual-poetic embedding model, annotators were requested to give a 0-10 scale score to a poem given an image considering their relevance in case of content, emotion and scene.

\textbf{Task2}: this task aims to compare the generated poems by different methods (four baseline methods and our four model settings) for one image on different aspects. Given an image, the annotators were asked to give ratings to a poem on a 0-10 scale with respect to four criteria: relevance (to the image), coherence (whether the poem is coherent across lines), imaginativeness (how much imaginative and creative the poem is for the given image) and overall impression.

\textbf{Task3}: Turing test was conducted by asking annotators to select human written poem from mixed human written and generated poems. Note that Turing test was implemented in two settings, i.e., with and without images as references.

For each task, we have randomly picked up 1K images and each task is assigned to three assessors. As poem is a form of literature, we also ask 30 annotators whose majors are related to English literature (among which ten annotations are English natives) as \textit{expert users} to do the Turing test.

\subsection{Training Details}

In the deep coupled visual-poetic embedding model, we use $D=4,096$-dimension ``fc7'' features for each CNN. Object features are extracted from VGG-16 \cite{simonyan2014very} trained on ImageNet \cite{russakovsky2015imagenet}, scene features from Place205-VGGNet model \cite{wang2015places205}, and sentiment features from sentiment model\cite{wang2016beyond}.

To better extract visual feature for poetic symbols, we first get nouns, verbs and adjectives with at least five frequency in UniM-Poem dataset. Then we manually picked adjectives and verbs for sentiment (including 328 labels), nouns for object (including 604 labels) and scenes (including 125 labels). As for poem features, we extract a combined skip-thought vector with $M=2,048$-dimension (in which each $1,024$-dimension represents for uni-direction and bi-direction, respectively) for each sentence, and finally we get poem features by mean pooling. And the margin $\alpha$ is set to $0.2$ based on empirical experiments in \cite{kiros2014unifying}. We randomly select 127 poems as unpaired poems for an image and used them as contrastive poems ($\mathbf{m}_k$ and $\mathbf{x}_k$ in Eq.~\eqref{eq:embedding}), and we re-sample them in each epoch.
Before adversarial training, we pre-train a generator based on image captioning method \cite{vinyals2015show} which can provide a better policy initialization for generator.
We empirically set the tradeoff parameter $\lambda=0.8$ by conducting a comparable observation on automatic evaluation results from $0.1$ to $0.9$.

\subsection{Evaluations}

\begin{table}[h]\small
\vspace{-2mm}
\small
\begin{center}
\begin{tabular}{c|c|c|c}
\hline
 & Ground-Truth & VPE w/o FT &VPE w/ FT\\
\hline\hline
Relevance& 7.22  & 5.82 & 6.32\\
\hline
\end{tabular}
\end{center}
\caption{\small{Average score of relevance to images for three types of human written poems on 0-10 scale (0-irrelevant, 10-relevant). One-way ANOVA revealed that evaluation on these poems is statistically significant ($F (2, 9) = 130.58, p < 1e-10)$.}}
\label{table:relevance}
\vspace{-7mm}
\end{table}

\textbf{Retrieved Poems.} We compare three kinds of poems considering their relevance to images: ground-truth poems, poems retrieved with VPE and image features before fine-tuning (VPE w/o FT), and poems retrieved with VPE and fine-tuned image features (VPE w/ FT). Table \ref{table:relevance} shows a comparison on a scale of 0-10 (0 means irrelevant and 10 means the most relevant). We can see that by using the proposed visual-poetic embedding model, the retrieved poems can achieve a relevance score above the average score (i.e., the score of five). And image features fine-tuned with poetic symbols can improve the relevance significantly.

\begin{table*}[t]\small
\begin{center}
\begin{tabular}{cccccccccc}
\hline
Method & Relevance &Novelty-2&Novelty-3  & BLEU-1&BLEU-2 &BLEU-3 &Overall\\
\hline\hline
Show and Tell (1CNN)\cite{vinyals2015show} & 1.79   & 43.66 & 76.76&11.88&3.35&0.76 &14.40\\ \hline
Show and Tell (3CNNs)\cite{vinyals2015show} & 1.91  & 48.09 & 81.37&12.64&3.34&0.8 &34.34\\ \hline
SeqGAN\cite{yu2017seqgan} & 2.03 & 47.52 & 82.32&13.40&3.72&0.76 &44.95\\ \hline
Regions-Hierarchical\cite{krause2016hierarchical} &1.81  & 46.75&79.90& 11.64&2.5&0.67 &8.01\\ \hline
\hline
I2P-GAN w/o discriminator & 1.94  &45.25 &80.13 & 13.35&3.69 &0.88 &41.86\\ \hline
I2P-GAN w/ $\mathrm{D_m}$  & 2.07 &43.37 &78.98 & \textbf{15.15}&\textbf{4.13} &\textbf{1.02} &63.00\\ \hline
I2P-GAN w/ $\mathrm{D_p}$ & 1.90  &\textbf{60.66} &\textbf{89.74} &12.91 &3.05 &0.72& 51.35 \\ \hline
I2P-GAN & \textbf{2.25}& 54.32&85.37& 14.25&3.84&0.94&\textbf{77.23}\\
\hline
\end{tabular}
\end{center}
\caption{\small{Automatic evaluation. Note that BLEU scores are computed in comparison with human-annotated ground-truth poems (one poem for one image). Overall score is computed as an average of three metrics after normalization (Eq.~\eqref{metric:normal}). All scores are reported as percentage (\%).}}
\label{table:autoresult}
\vspace{-8mm}
\end{table*}

\textbf{Generated Poems.}
Table \ref{table:autoresult} exhibits the automatic evaluation results of the proposed model with four settings, as well as the four baselines proposed in previous works.
Comparing results of caption model with one CNN and three CNNs, we can see that multi-CNN can actually help to generate poems that are more relevant to images.
Regions-Hierarchical model emphasizes more on the topic coherence between sentences while many human written poems will cover several topics or use different symbols for one topic.
SeqGAN shows the advantage of applying adversarial training for poem generation compared with only caption models with only CNN-RNN while lacking of generating novel concepts in poems.
Better performance of our pre-trained model with VPE than caption model demonstrates the effectiveness of VPE in extracting poetic features from images for better poem generation.
We can see that our three models outperform in most of the metrics with each one performs better at one aspect.
The model with only multi-modal discriminator (\textbf{I2P-GAN w/ $\mathbf{D_m}$}) will guide the model to generate poems towards ground-truth poems, thus it results in the highest BLEU scores that emphasize the similarity of n-grams in a translative way.
Poem-style discriminator ($D_p$) is designed to guide the generated poem to be more poetic in language style, and the highest novelty score of \textbf{I2P-GAN w/ $\mathbf{D_m}$} shows that $D_p$ helps to provide more novel and imaginative words to the generated poem.
Overall, \textbf{I2P-GAN} combines the advantages of both discriminators with a rational intermediate score regarding BLEU and novelty while still outperforms compared with other generation models. Moreover, our model with both discriminators can generate poems that have highest relevance on our embedding relevance metric.

\begin{table}[h]\small
\begin{center}
\begin{tabular}{ccccc}
\hline
Method & Rel & Col & Imag & Overall\\
\hline\hline
Show and Tell (1CNN)\cite{vinyals2015show} & 6.31   & 6.52 & 6.57&6.67\\ \hline
Show and Tell (3CNNs)\cite{vinyals2015show} & 6.41  & 6.59 & 6.63&6.75\\ \hline
SeqGAN\cite{yu2017seqgan} & 6.13 & 6.43 & 6.50&6.63\\ \hline
Regions-Hierarchical\cite{krause2016hierarchical} &6.35  & 6.54&6.63& 6.78\\ \hline
\hline
I2P-GAN w/o discriminator & 6.44  &6.64 &6.77 & 6.85\\ \hline
I2P-GAN w/ $\mathrm{D_m}$  & 6.59 &6.83 &6.94 & 7.06\\ \hline
I2P-GAN w/ $\mathrm{D_p}$ & 6.53  &6.75 &6.80 &6.93 \\ \hline
I2P-GAN & \textbf{6.83}& \textbf{6.95}&\textbf{7.05}& \textbf{7.18}\\ \hline
\hline
Ground-Truth & 7.10 & 7.26 & 7.23 & 7.37 \\
\hline
\end{tabular}
\end{center}
\caption{\small{Human evaluation results of six methods on four criteria: relevance (Rel), coherence (Col), imaginativeness (Imag) and Overall. All criteria are evaluated on 0-10 scale (0-bad, 10-good).}}
\label{table:humaneval}
\vspace{-8mm}
\end{table}

Comparison of human evaluation results are shown in Table \ref{table:humaneval}. Different from automatic evaluation results where Regions-Hierarchical performs not well, it gets a slightly better result than caption model for the reason that sentences all about the same topic tend to gain better impressions from users. Our three models outperform the other four baseline methods on all metrics. Two discriminators promote human-level comprehension towards poems compared with pre-trained model. The model with two discriminators has generated better poems from images in terms of relevance, coherence and imaginativeness.
Fig.~(\ref{fig:comparedexample}) shows one example of poems generated with three baselines and our methods for a given image.
More examples generated by our approach can be referred in Fig.~(\ref{fig:generatedexample}).

\begin{table}[h]
\small
\begin{center}
\begin{tabular}{cccc}
\hline
Data&Users & Ground-Truth  &Generated\\
\hline\hline
\multirow{2}*{Poem w/ Image} &AMT&  0.51  & 0.49 \\ \cline{2-4}
&Expert& 0.60   & 0.40\\ \hline \hline
\multirow{2}*{Poem w/o Image} &AMT&  0.55  & 0.45 \\ \cline{2-4}
&Expert& 0.57   & 0.43\\ \hline
\end{tabular}
\end{center}
\caption{\small{Accuracy of Turing test on AMT users and expert users on poems with and without images.}}
\label{table:turing}
\vspace{-7mm}
\end{table}

\textbf{Turing Test.}
For the Turing test of annotators in AMT, we have hired 548 workers with 10.9 tasks for each one on average. For experts, 15 people were asked to judge human written poems with images and another 15 annotators were asked to do test with only poems. Each one is assigned with 20 images and in total we have 600 tasks conducted by expert users. Table \ref{table:turing} shows the probability of different poems being selected as human-written poems for an given image. As we can see, the generated poems have caused a competitive confusion to both ordinary annotators and experts though experts can figure out the accurate one better than ordinary people. One interesting observation comes from that experts are better at figuring out correct ones with images while AMT workers do better with only poems.

\section{Conclusion}

As the frontal work of poetry (English free verse) generation from images, we propose a novel approach to model the problem by incorporating deep coupled visual-poetic embedding model and RNN based adversarial training with multi-discriminators as rewards for policy gradient. Furthermore, we introduce the first image and poem pair dataset (MultiM-Poem) and a large poem corpus (UniM-Poem) to enhance researches on poem generation, especially from images.
Extensive experiments demonstrated that our embedding model can approximately learn a rational visual-poetic embedding space. Objective and subjective evaluation results demonstrated the effectiveness of our poem generation model.

\begin{acks}
This work was supported by JSPS KAKENHI Grant Number 18H03244 and 18H03243.
The authors would like to thank Dr. Tao Mei for his insightful discussions on this work when he was a senior research manager at Microsoft Research, and thank Yuanchun Xu, Guang Zhou, Hongyuan Zhu, Yang Xiang and Yu Shi for promoting the technology transfer of this research to Microsoft XiaoIce.
\end{acks}

\bibliographystyle{ACM-Reference-Format}
\balance
\bibliography{img2poem}


\begin{thebibliography}{41}


\ifx \showCODEN    \undefined \def \showCODEN     #1{\unskip}     \fi
\ifx \showDOI      \undefined \def \showDOI       #1{#1}\fi
\ifx \showISBNx    \undefined \def \showISBNx     #1{\unskip}     \fi
\ifx \showISBNxiii \undefined \def \showISBNxiii  #1{\unskip}     \fi
\ifx \showISSN     \undefined \def \showISSN      #1{\unskip}     \fi
\ifx \showLCCN     \undefined \def \showLCCN      #1{\unskip}     \fi
\ifx \shownote     \undefined \def \shownote      #1{#1}          \fi
\ifx \showarticletitle \undefined \def \showarticletitle #1{#1}   \fi
\ifx \showURL      \undefined \def \showURL       {\relax}        \fi
\providecommand\bibfield[2]{#2}
\providecommand\bibinfo[2]{#2}
\providecommand\natexlab[1]{#1}
\providecommand\showeprint[2][]{arXiv:#2}

\bibitem[\protect\citeauthoryear{Chen, Liao, Chuang, Hsu, Fu, and Sun}{Chen
  et~al\mbox{.}}{2017}]%
        {chen2017show}
\bibfield{author}{\bibinfo{person}{Tseng-Hung Chen}, \bibinfo{person}{Yuan-Hong
  Liao}, \bibinfo{person}{Ching-Yao Chuang}, \bibinfo{person}{Wan-Ting Hsu},
  \bibinfo{person}{Jianlong Fu}, {and} \bibinfo{person}{Min Sun}.}
  \bibinfo{year}{2017}\natexlab{}.
\newblock \showarticletitle{Show, Adapt and Tell: Adversarial Training of
  Cross-domain Image Captioner}.
\newblock \bibinfo{journal}{\emph{ICCV}} (\bibinfo{year}{2017}),
  \bibinfo{pages}{521--530}.
\newblock


\bibitem[\protect\citeauthoryear{Chen and Lawrence~Zitnick}{Chen and
  Lawrence~Zitnick}{2015}]%
        {chen2015mind}
\bibfield{author}{\bibinfo{person}{Xinlei Chen} {and} \bibinfo{person}{C
  Lawrence~Zitnick}.} \bibinfo{year}{2015}\natexlab{}.
\newblock \showarticletitle{Mind's eye: A recurrent visual representation for
  image caption generation}. In \bibinfo{booktitle}{\emph{CVPR}}.
  \bibinfo{pages}{2422--2431}.
\newblock


\bibitem[\protect\citeauthoryear{Chung, Gulcehre, Cho, and Bengio}{Chung
  et~al\mbox{.}}{2014}]%
        {chung2014empirical}
\bibfield{author}{\bibinfo{person}{Junyoung Chung}, \bibinfo{person}{Caglar
  Gulcehre}, \bibinfo{person}{KyungHyun Cho}, {and} \bibinfo{person}{Yoshua
  Bengio}.} \bibinfo{year}{2014}\natexlab{}.
\newblock \showarticletitle{Empirical evaluation of gated recurrent neural
  networks on sequence modeling}.
\newblock \bibinfo{journal}{\emph{NIPS 2014 Workshop on Deep Learning}}
  (\bibinfo{year}{2014}).
\newblock


\bibitem[\protect\citeauthoryear{Fang, Gupta, Iandola, Srivastava, Deng,
  Doll{\'a}r, Gao, He, Mitchell, Platt, et~al\mbox{.}}{Fang
  et~al\mbox{.}}{2015}]%
        {fang2015captions}
\bibfield{author}{\bibinfo{person}{Hao Fang}, \bibinfo{person}{Saurabh Gupta},
  \bibinfo{person}{Forrest Iandola}, \bibinfo{person}{Rupesh~K Srivastava},
  \bibinfo{person}{Li Deng}, \bibinfo{person}{Piotr Doll{\'a}r},
  \bibinfo{person}{Jianfeng Gao}, \bibinfo{person}{Xiaodong He},
  \bibinfo{person}{Margaret Mitchell}, \bibinfo{person}{John~C Platt},
  {et~al\mbox{.}}} \bibinfo{year}{2015}\natexlab{}.
\newblock \showarticletitle{From captions to visual concepts and back}. In
  \bibinfo{booktitle}{\emph{CVPR}}. \bibinfo{pages}{1473--1482}.
\newblock


\bibitem[\protect\citeauthoryear{Farhadi, Hejrati, Sadeghi, Young, Rashtchian,
  Hockenmaier, and Forsyth}{Farhadi et~al\mbox{.}}{2010}]%
        {farhadi2010every}
\bibfield{author}{\bibinfo{person}{Ali Farhadi}, \bibinfo{person}{Mohsen
  Hejrati}, \bibinfo{person}{Mohammad~Amin Sadeghi}, \bibinfo{person}{Peter
  Young}, \bibinfo{person}{Cyrus Rashtchian}, \bibinfo{person}{Julia
  Hockenmaier}, {and} \bibinfo{person}{David Forsyth}.}
  \bibinfo{year}{2010}\natexlab{}.
\newblock \showarticletitle{Every picture tells a story: Generating sentences
  from images}. In \bibinfo{booktitle}{\emph{ECCV}}. \bibinfo{pages}{15--29}.
\newblock


\bibitem[\protect\citeauthoryear{Frome, Corrado, Shlens, Bengio, Dean, Mikolov,
  et~al\mbox{.}}{Frome et~al\mbox{.}}{2013}]%
        {frome2013devise}
\bibfield{author}{\bibinfo{person}{Andrea Frome}, \bibinfo{person}{Greg~S
  Corrado}, \bibinfo{person}{Jon Shlens}, \bibinfo{person}{Samy Bengio},
  \bibinfo{person}{Jeff Dean}, \bibinfo{person}{Tomas Mikolov},
  {et~al\mbox{.}}} \bibinfo{year}{2013}\natexlab{}.
\newblock \showarticletitle{Devise: A deep visual-semantic embedding model}. In
  \bibinfo{booktitle}{\emph{NIPS}}. \bibinfo{pages}{2121--2129}.
\newblock


\bibitem[\protect\citeauthoryear{Fu, Zheng, and Mei}{Fu et~al\mbox{.}}{2017}]%
        {LookCloser}
\bibfield{author}{\bibinfo{person}{Jianlong Fu}, \bibinfo{person}{Heliang
  Zheng}, {and} \bibinfo{person}{Tao Mei}.} \bibinfo{year}{2017}\natexlab{}.
\newblock \showarticletitle{Look Closer to See Better: Recurrent Attention
  Convolutional Neural Network for Fine-grained Image Recognition}. In
  \bibinfo{booktitle}{\emph{CVPR}}. \bibinfo{pages}{4438--4446}.
\newblock


\bibitem[\protect\citeauthoryear{Ghazvininejad, Shi, Choi, and
  Knight}{Ghazvininejad et~al\mbox{.}}{2016}]%
        {ghazvininejad2016generating}
\bibfield{author}{\bibinfo{person}{Marjan Ghazvininejad}, \bibinfo{person}{Xing
  Shi}, \bibinfo{person}{Yejin Choi}, {and} \bibinfo{person}{Kevin Knight}.}
  \bibinfo{year}{2016}\natexlab{}.
\newblock \showarticletitle{Generating Topical Poetry.}. In
  \bibinfo{booktitle}{\emph{EMNLP}}. \bibinfo{pages}{1183--1191}.
\newblock


\bibitem[\protect\citeauthoryear{Ghazvininejad, Shi, Priyadarshi, and
  Knight}{Ghazvininejad et~al\mbox{.}}{2017}]%
        {ghazvininejad2017hafez}
\bibfield{author}{\bibinfo{person}{Marjan Ghazvininejad}, \bibinfo{person}{Xing
  Shi}, \bibinfo{person}{Jay Priyadarshi}, {and} \bibinfo{person}{Kevin
  Knight}.} \bibinfo{year}{2017}\natexlab{}.
\newblock \showarticletitle{Hafez: an Interactive Poetry Generation System}.
\newblock \bibinfo{journal}{\emph{ACL}} (\bibinfo{year}{2017}),
  \bibinfo{pages}{43--48}.
\newblock


\bibitem[\protect\citeauthoryear{Goodfellow, Pouget-Abadie, Mirza, Xu,
  Warde-Farley, Ozair, Courville, and Bengio}{Goodfellow et~al\mbox{.}}{2014}]%
        {goodfellow2014generative}
\bibfield{author}{\bibinfo{person}{Ian Goodfellow}, \bibinfo{person}{Jean
  Pouget-Abadie}, \bibinfo{person}{Mehdi Mirza}, \bibinfo{person}{Bing Xu},
  \bibinfo{person}{David Warde-Farley}, \bibinfo{person}{Sherjil Ozair},
  \bibinfo{person}{Aaron Courville}, {and} \bibinfo{person}{Yoshua Bengio}.}
  \bibinfo{year}{2014}\natexlab{}.
\newblock \showarticletitle{Generative adversarial nets}. In
  \bibinfo{booktitle}{\emph{NIPS}}. \bibinfo{pages}{2672--2680}.
\newblock


\bibitem[\protect\citeauthoryear{He, Zhou, and Jiang}{He et~al\mbox{.}}{2012}]%
        {he2012generating}
\bibfield{author}{\bibinfo{person}{Jing He}, \bibinfo{person}{Ming Zhou}, {and}
  \bibinfo{person}{Long Jiang}.} \bibinfo{year}{2012}\natexlab{}.
\newblock \showarticletitle{Generating Chinese Classical Poems with Statistical
  Machine Translation Models.}. In \bibinfo{booktitle}{\emph{AAAI}}.
\newblock


\bibitem[\protect\citeauthoryear{Hopkins and Kiela}{Hopkins and Kiela}{2017}]%
        {hopkins2017automatically}
\bibfield{author}{\bibinfo{person}{Jack Hopkins} {and} \bibinfo{person}{Douwe
  Kiela}.} \bibinfo{year}{2017}\natexlab{}.
\newblock \showarticletitle{Automatically Generating Rhythmic Verse with Neural
  Networks}. In \bibinfo{booktitle}{\emph{ACL}}, Vol.~\bibinfo{volume}{1}.
  \bibinfo{pages}{168--178}.
\newblock


\bibitem[\protect\citeauthoryear{Jiang and Zhou}{Jiang and Zhou}{2008}]%
        {jiang2008generating}
\bibfield{author}{\bibinfo{person}{Long Jiang} {and} \bibinfo{person}{Ming
  Zhou}.} \bibinfo{year}{2008}\natexlab{}.
\newblock \showarticletitle{Generating Chinese couplets using a statistical MT
  approach}. In \bibinfo{booktitle}{\emph{COLING}}. \bibinfo{pages}{377--384}.
\newblock


\bibitem[\protect\citeauthoryear{Karpathy, Joulin, and Li}{Karpathy
  et~al\mbox{.}}{2014}]%
        {karpathy2014deep}
\bibfield{author}{\bibinfo{person}{Andrej Karpathy}, \bibinfo{person}{Armand
  Joulin}, {and} \bibinfo{person}{Fei Fei~F Li}.}
  \bibinfo{year}{2014}\natexlab{}.
\newblock \showarticletitle{Deep fragment embeddings for bidirectional image
  sentence mapping}. In \bibinfo{booktitle}{\emph{NIPS}}.
  \bibinfo{pages}{1889--1897}.
\newblock


\bibitem[\protect\citeauthoryear{Kiros, Salakhutdinov, and Zemel}{Kiros
  et~al\mbox{.}}{2014}]%
        {kiros2014unifying}
\bibfield{author}{\bibinfo{person}{Ryan Kiros}, \bibinfo{person}{Ruslan
  Salakhutdinov}, {and} \bibinfo{person}{Richard~S Zemel}.}
  \bibinfo{year}{2014}\natexlab{}.
\newblock \showarticletitle{Unifying visual-semantic embeddings with multimodal
  neural language models}.
\newblock \bibinfo{journal}{\emph{arXiv preprint arXiv:1411.2539}}
  (\bibinfo{year}{2014}).
\newblock


\bibitem[\protect\citeauthoryear{Kiros, Zhu, Salakhutdinov, Zemel, Urtasun,
  Torralba, and Fidler}{Kiros et~al\mbox{.}}{2015}]%
        {kiros2015skip}
\bibfield{author}{\bibinfo{person}{Ryan Kiros}, \bibinfo{person}{Yukun Zhu},
  \bibinfo{person}{Ruslan~R Salakhutdinov}, \bibinfo{person}{Richard Zemel},
  \bibinfo{person}{Raquel Urtasun}, \bibinfo{person}{Antonio Torralba}, {and}
  \bibinfo{person}{Sanja Fidler}.} \bibinfo{year}{2015}\natexlab{}.
\newblock \showarticletitle{Skip-thought vectors}. In
  \bibinfo{booktitle}{\emph{NIPS}}. \bibinfo{pages}{3294--3302}.
\newblock


\bibitem[\protect\citeauthoryear{Krause, Johnson, Krishna, and Fei-Fei}{Krause
  et~al\mbox{.}}{2017}]%
        {krause2016hierarchical}
\bibfield{author}{\bibinfo{person}{Jonathan Krause}, \bibinfo{person}{Justin
  Johnson}, \bibinfo{person}{Ranjay Krishna}, {and} \bibinfo{person}{Li
  Fei-Fei}.} \bibinfo{year}{2017}\natexlab{}.
\newblock \showarticletitle{A hierarchical approach for generating descriptive
  image paragraphs}.
\newblock \bibinfo{journal}{\emph{CVPR}} (\bibinfo{year}{2017}),
  \bibinfo{pages}{3337--3345}.
\newblock


\bibitem[\protect\citeauthoryear{Kulkarni, Premraj, Dhar, Li, Choi, Berg, and
  Berg}{Kulkarni et~al\mbox{.}}{2011}]%
        {kulkarni2011baby}
\bibfield{author}{\bibinfo{person}{Girish Kulkarni}, \bibinfo{person}{Visruth
  Premraj}, \bibinfo{person}{Sagnik Dhar}, \bibinfo{person}{Siming Li},
  \bibinfo{person}{Yejin Choi}, \bibinfo{person}{Alexander~C Berg}, {and}
  \bibinfo{person}{Tamara~L Berg}.} \bibinfo{year}{2011}\natexlab{}.
\newblock \showarticletitle{Baby talk: Understanding and generating image
  descriptions}. In \bibinfo{booktitle}{\emph{CVPR}}.
  \bibinfo{pages}{1601--1608}.
\newblock


\bibitem[\protect\citeauthoryear{Liu, Fu, Mei, and Chen}{Liu
  et~al\mbox{.}}{2017}]%
        {liu2017let}
\bibfield{author}{\bibinfo{person}{Yu Liu}, \bibinfo{person}{Jianlong Fu},
  \bibinfo{person}{Tao Mei}, {and} \bibinfo{person}{Chang~Wen Chen}.}
  \bibinfo{year}{2017}\natexlab{}.
\newblock \showarticletitle{Let Your Photos Talk: Generating Narrative
  Paragraph for Photo Stream via Bidirectional Attention Recurrent Neural
  Networks}. In \bibinfo{booktitle}{\emph{AAAI}}.
\newblock


\bibitem[\protect\citeauthoryear{Manurung}{Manurung}{1999}]%
        {manurung1999chart}
\bibfield{author}{\bibinfo{person}{Hisar~Maruli Manurung}.}
  \bibinfo{year}{1999}\natexlab{}.
\newblock \showarticletitle{A chart generator for rhythm patterned text}. In
  \bibinfo{booktitle}{\emph{Proceedings of the First International Workshop on
  Literature in Cognition and Computer}}. \bibinfo{pages}{15--19}.
\newblock


\bibitem[\protect\citeauthoryear{Oliveira}{Oliveira}{2009}]%
        {oliveira2009automatic}
\bibfield{author}{\bibinfo{person}{Hugo Oliveira}.}
  \bibinfo{year}{2009}\natexlab{}.
\newblock \showarticletitle{Automatic generation of poetry: an overview}.
\newblock \bibinfo{journal}{\emph{Universidade de Coimbra}}
  (\bibinfo{year}{2009}).
\newblock


\bibitem[\protect\citeauthoryear{Oliveira}{Oliveira}{2012}]%
        {oliveira2012poetryme}
\bibfield{author}{\bibinfo{person}{Hugo~Gon{\c{c}}alo Oliveira}.}
  \bibinfo{year}{2012}\natexlab{}.
\newblock \showarticletitle{PoeTryMe: a versatile platform for poetry
  generation}.
\newblock \bibinfo{journal}{\emph{Computational Creativity, Concept Invention,
  and General Intelligence}}  \bibinfo{volume}{1} (\bibinfo{year}{2012}),
  \bibinfo{pages}{21}.
\newblock


\bibitem[\protect\citeauthoryear{Papineni, Roukos, Ward, and Zhu}{Papineni
  et~al\mbox{.}}{2002}]%
        {papineni2002bleu}
\bibfield{author}{\bibinfo{person}{Kishore Papineni}, \bibinfo{person}{Salim
  Roukos}, \bibinfo{person}{Todd Ward}, {and} \bibinfo{person}{Wei-Jing Zhu}.}
  \bibinfo{year}{2002}\natexlab{}.
\newblock \showarticletitle{BLEU: a method for automatic evaluation of machine
  translation}. In \bibinfo{booktitle}{\emph{ACL}}. \bibinfo{pages}{311--318}.
\newblock


\bibitem[\protect\citeauthoryear{Park and Kim}{Park and Kim}{2015}]%
        {park2015expressing}
\bibfield{author}{\bibinfo{person}{Cesc~C Park} {and} \bibinfo{person}{Gunhee
  Kim}.} \bibinfo{year}{2015}\natexlab{}.
\newblock \showarticletitle{Expressing an image stream with a sequence of
  natural sentences}. In \bibinfo{booktitle}{\emph{NIPS}}.
  \bibinfo{pages}{73--81}.
\newblock


\bibitem[\protect\citeauthoryear{Rennie, Marcheret, Mroueh, Ross, and
  Goel}{Rennie et~al\mbox{.}}{2017}]%
        {rennie2017self}
\bibfield{author}{\bibinfo{person}{Steven~J Rennie}, \bibinfo{person}{Etienne
  Marcheret}, \bibinfo{person}{Youssef Mroueh}, \bibinfo{person}{Jarret Ross},
  {and} \bibinfo{person}{Vaibhava Goel}.} \bibinfo{year}{2017}\natexlab{}.
\newblock \showarticletitle{Self-critical sequence training for image
  captioning}. In \bibinfo{booktitle}{\emph{CVPR}}, Vol.~\bibinfo{volume}{1}.
  \bibinfo{pages}{3}.
\newblock


\bibitem[\protect\citeauthoryear{Russakovsky, Deng, Su, Krause, Satheesh, Ma,
  Huang, Karpathy, Khosla, Bernstein, et~al\mbox{.}}{Russakovsky
  et~al\mbox{.}}{2015}]%
        {russakovsky2015imagenet}
\bibfield{author}{\bibinfo{person}{Olga Russakovsky}, \bibinfo{person}{Jia
  Deng}, \bibinfo{person}{Hao Su}, \bibinfo{person}{Jonathan Krause},
  \bibinfo{person}{Sanjeev Satheesh}, \bibinfo{person}{Sean Ma},
  \bibinfo{person}{Zhiheng Huang}, \bibinfo{person}{Andrej Karpathy},
  \bibinfo{person}{Aditya Khosla}, \bibinfo{person}{Michael Bernstein},
  {et~al\mbox{.}}} \bibinfo{year}{2015}\natexlab{}.
\newblock \showarticletitle{Imagenet large scale visual recognition challenge}.
\newblock \bibinfo{journal}{\emph{IJCV}} \bibinfo{volume}{115},
  \bibinfo{number}{3} (\bibinfo{year}{2015}), \bibinfo{pages}{211--252}.
\newblock


\bibitem[\protect\citeauthoryear{Simonyan and Zisserman}{Simonyan and
  Zisserman}{2014}]%
        {simonyan2014very}
\bibfield{author}{\bibinfo{person}{Karen Simonyan} {and}
  \bibinfo{person}{Andrew Zisserman}.} \bibinfo{year}{2014}\natexlab{}.
\newblock \showarticletitle{Very deep convolutional networks for large-scale
  image recognition}.
\newblock \bibinfo{journal}{\emph{arXiv preprint arXiv:1409.1556}}
  (\bibinfo{year}{2014}).
\newblock


\bibitem[\protect\citeauthoryear{Toutanova, Klein, Manning, and
  Singer}{Toutanova et~al\mbox{.}}{2003}]%
        {toutanova2003feature}
\bibfield{author}{\bibinfo{person}{Kristina Toutanova}, \bibinfo{person}{Dan
  Klein}, \bibinfo{person}{Christopher~D Manning}, {and} \bibinfo{person}{Yoram
  Singer}.} \bibinfo{year}{2003}\natexlab{}.
\newblock \showarticletitle{Feature-rich part-of-speech tagging with a cyclic
  dependency network}. In \bibinfo{booktitle}{\emph{HLT-NAACL}}.
  \bibinfo{pages}{173--180}.
\newblock


\bibitem[\protect\citeauthoryear{Vinyals, Toshev, Bengio, and Erhan}{Vinyals
  et~al\mbox{.}}{2015}]%
        {vinyals2015show}
\bibfield{author}{\bibinfo{person}{Oriol Vinyals}, \bibinfo{person}{Alexander
  Toshev}, \bibinfo{person}{Samy Bengio}, {and} \bibinfo{person}{Dumitru
  Erhan}.} \bibinfo{year}{2015}\natexlab{}.
\newblock \showarticletitle{Show and tell: A neural image caption generator}.
  In \bibinfo{booktitle}{\emph{CVPR}}. \bibinfo{pages}{3156--3164}.
\newblock


\bibitem[\protect\citeauthoryear{Wang, Fu, Xu, and Mei}{Wang
  et~al\mbox{.}}{2016}]%
        {wang2016beyond}
\bibfield{author}{\bibinfo{person}{Jingwen Wang}, \bibinfo{person}{Jianlong
  Fu}, \bibinfo{person}{Yong Xu}, {and} \bibinfo{person}{Tao Mei}.}
  \bibinfo{year}{2016}\natexlab{}.
\newblock \showarticletitle{Beyond Object Recognition: Visual Sentiment
  Analysis with Deep Coupled Adjective and Noun Neural Networks.}. In
  \bibinfo{booktitle}{\emph{IJCAI}}. \bibinfo{pages}{3484--3490}.
\newblock


\bibitem[\protect\citeauthoryear{Wang, Guo, Huang, and Qiao}{Wang
  et~al\mbox{.}}{2015}]%
        {wang2015places205}
\bibfield{author}{\bibinfo{person}{Limin Wang}, \bibinfo{person}{Sheng Guo},
  \bibinfo{person}{Weilin Huang}, {and} \bibinfo{person}{Yu Qiao}.}
  \bibinfo{year}{2015}\natexlab{}.
\newblock \showarticletitle{Places205-vggnet models for scene recognition}.
\newblock \bibinfo{journal}{\emph{arXiv preprint arXiv:1508.01667}}
  (\bibinfo{year}{2015}).
\newblock


\bibitem[\protect\citeauthoryear{Williams}{Williams}{1992}]%
        {williams1992simple}
\bibfield{author}{\bibinfo{person}{Ronald~J Williams}.}
  \bibinfo{year}{1992}\natexlab{}.
\newblock \showarticletitle{Simple statistical gradient-following algorithms
  for connectionist reinforcement learning}.
\newblock \bibinfo{journal}{\emph{Machine learning}} \bibinfo{volume}{8},
  \bibinfo{number}{3-4} (\bibinfo{year}{1992}), \bibinfo{pages}{229--256}.
\newblock


\bibitem[\protect\citeauthoryear{Xu, Jiang, Qin, Wang, and Du}{Xu
  et~al\mbox{.}}{2018}]%
        {xu2018images}
\bibfield{author}{\bibinfo{person}{Linli Xu}, \bibinfo{person}{Liang Jiang},
  \bibinfo{person}{Chuan Qin}, \bibinfo{person}{Zhe Wang}, {and}
  \bibinfo{person}{Dongfang Du}.} \bibinfo{year}{2018}\natexlab{}.
\newblock \showarticletitle{How Images Inspire Poems: Generating Classical
  Chinese Poetry from Images with Memory Networks}. In
  \bibinfo{booktitle}{\emph{AAAI}}.
\newblock


\bibitem[\protect\citeauthoryear{Xu, Liu, Wang, Chengjie, Wang, Wang, and
  Qi}{Xu et~al\mbox{.}}{2017}]%
        {xu2017neural}
\bibfield{author}{\bibinfo{person}{Zhen Xu}, \bibinfo{person}{Bingquan Liu},
  \bibinfo{person}{Baoxun Wang}, \bibinfo{person}{SUN Chengjie},
  \bibinfo{person}{Xiaolong Wang}, \bibinfo{person}{Zhuoran Wang}, {and}
  \bibinfo{person}{Chao Qi}.} \bibinfo{year}{2017}\natexlab{}.
\newblock \showarticletitle{Neural Response Generation via GAN with an
  Approximate Embedding Layer}. In \bibinfo{booktitle}{\emph{EMNLP}}.
  \bibinfo{pages}{628--637}.
\newblock


\bibitem[\protect\citeauthoryear{Yan, Jiang, Lapata, Lin, Lv, and Li}{Yan
  et~al\mbox{.}}{2013}]%
        {yan2013poet}
\bibfield{author}{\bibinfo{person}{Rui Yan}, \bibinfo{person}{Han Jiang},
  \bibinfo{person}{Mirella Lapata}, \bibinfo{person}{Shou-De Lin},
  \bibinfo{person}{Xueqiang Lv}, {and} \bibinfo{person}{Xiaoming Li}.}
  \bibinfo{year}{2013}\natexlab{}.
\newblock \showarticletitle{i, Poet: Automatic Chinese Poetry Composition
  through a Generative Summarization Framework under Constrained
  Optimization.}. In \bibinfo{booktitle}{\emph{IJCAI}}.
  \bibinfo{pages}{2197--2203}.
\newblock


\bibitem[\protect\citeauthoryear{Yao, Pan, Li, Qiu, and Mei}{Yao
  et~al\mbox{.}}{2017}]%
        {yao2017boosting}
\bibfield{author}{\bibinfo{person}{Ting Yao}, \bibinfo{person}{Yingwei Pan},
  \bibinfo{person}{Yehao Li}, \bibinfo{person}{Zhaofan Qiu}, {and}
  \bibinfo{person}{Tao Mei}.} \bibinfo{year}{2017}\natexlab{}.
\newblock \showarticletitle{Boosting image captioning with attributes}. In
  \bibinfo{booktitle}{\emph{IEEE International Conference on Computer Vision,
  ICCV}}. \bibinfo{pages}{22--29}.
\newblock


\bibitem[\protect\citeauthoryear{Yi, Li, and Sun}{Yi et~al\mbox{.}}{2017}]%
        {yi2017generating}
\bibfield{author}{\bibinfo{person}{Xiaoyuan Yi}, \bibinfo{person}{Ruoyu Li},
  {and} \bibinfo{person}{Maosong Sun}.} \bibinfo{year}{2017}\natexlab{}.
\newblock \showarticletitle{Generating Chinese Classical Poems with RNN
  Encoder-Decoder}.
\newblock In \bibinfo{booktitle}{\emph{Chinese Computational Linguistics and
  Natural Language Processing Based on Naturally Annotated Big Data}}.
  \bibinfo{pages}{211--223}.
\newblock


\bibitem[\protect\citeauthoryear{You, Jin, Wang, Fang, and Luo}{You
  et~al\mbox{.}}{2016}]%
        {you2016image}
\bibfield{author}{\bibinfo{person}{Quanzeng You}, \bibinfo{person}{Hailin Jin},
  \bibinfo{person}{Zhaowen Wang}, \bibinfo{person}{Chen Fang}, {and}
  \bibinfo{person}{Jiebo Luo}.} \bibinfo{year}{2016}\natexlab{}.
\newblock \showarticletitle{Image captioning with semantic attention}. In
  \bibinfo{booktitle}{\emph{CVPR}}. \bibinfo{pages}{4651--4659}.
\newblock


\bibitem[\protect\citeauthoryear{Yu, Zhang, Wang, and Yu}{Yu
  et~al\mbox{.}}{2017}]%
        {yu2017seqgan}
\bibfield{author}{\bibinfo{person}{Lantao Yu}, \bibinfo{person}{Weinan Zhang},
  \bibinfo{person}{Jun Wang}, {and} \bibinfo{person}{Yong Yu}.}
  \bibinfo{year}{2017}\natexlab{}.
\newblock \showarticletitle{SeqGAN: Sequence Generative Adversarial Nets with
  Policy Gradient.}. In \bibinfo{booktitle}{\emph{AAAI}}.
  \bibinfo{pages}{2852--2858}.
\newblock


\bibitem[\protect\citeauthoryear{Zaremba and Sutskever}{Zaremba and
  Sutskever}{2015}]%
        {zaremba2015reinforcement}
\bibfield{author}{\bibinfo{person}{Wojciech Zaremba} {and}
  \bibinfo{person}{Ilya Sutskever}.} \bibinfo{year}{2015}\natexlab{}.
\newblock \showarticletitle{Reinforcement Learning Neural Turing
  Machines-Revised}.
\newblock \bibinfo{journal}{\emph{arXiv preprint arXiv:1505.00521}}
  (\bibinfo{year}{2015}).
\newblock


\bibitem[\protect\citeauthoryear{Zhang and Lapata}{Zhang and Lapata}{2014}]%
        {zhang2014chinese}
\bibfield{author}{\bibinfo{person}{Xingxing Zhang} {and}
  \bibinfo{person}{Mirella Lapata}.} \bibinfo{year}{2014}\natexlab{}.
\newblock \showarticletitle{Chinese Poetry Generation with Recurrent Neural
  Networks.}. In \bibinfo{booktitle}{\emph{EMNLP}}. \bibinfo{pages}{670--680}.
\newblock


\end{thebibliography}

\end{document}